\documentclass{article}

\usepackage[preprint]{neurips_2024}

\usepackage[utf8]{inputenc} 
\usepackage[T1]{fontenc}    
\usepackage{url}            
\usepackage{booktabs}       
\usepackage{amsfonts}       
\usepackage{nicefrac}       
\usepackage{microtype}      

\usepackage{footnote}
\makesavenoteenv{tabular}
\makesavenoteenv{table}
\usepackage{amsmath}
\usepackage{xcolor}         
\usepackage{hyperref}

\usepackage{graphicx}%
\usepackage{multirow}%
\usepackage{amsmath,amssymb,amsfonts}%
\usepackage{amsthm}%
\usepackage{mathrsfs}%
\usepackage[title]{appendix}%
\usepackage{xcolor}%
\usepackage{textcomp}%
\usepackage{manyfoot}%
\usepackage{booktabs}%
\usepackage{algorithm}%
\usepackage{algorithmicx}%
\usepackage{algpseudocode}%
\usepackage{listings}%
\usepackage{float}
\usepackage{booktabs}
\usepackage{subcaption}
\usepackage{tabularx}
\usepackage{amssymb} 
\usepackage{graphicx}
\usepackage{array}
\usepackage{longtable}

\newcommand{\rheader}[1]{\rotatebox{30}{\parbox{2cm}{\centering #1}}}

\title{LRW-Persian: Lip‐reading in the Wild Dataset for Persian Language}

\author{%
  Zahra Taghizadeh \\
  Department of Mechanical Engineering \\
  Sharif University of Technology, Tehran, Iran \\ 
  \texttt{zahra.taghizade@epfl.ch} \\
  \And
  Mohammad Shahverdikondori \\
  Department of Mathematical Sciences  \\
  Sharif University of Technology, Tehran, Iran  \\
  \And
  Arian Noori\\
  Department of Computer Engineernig \\
  Sharif University of Technology, Tehran, Iran  \\
  \And 
  Alireza Dadgarnia \\
  Department of Mathematical Sciences \\
  Sharif University of Technology, Tehran, Iran \\
}

\begin{document}

\maketitle

\begin{abstract}
Lipreading has emerged as an increasingly important research area for developing robust speech recognition systems and assistive technologies for the hearing-impaired. However, non-English resources for visual speech recognition remain limited. We introduce \emph{LRW-Persian}, the largest in-the-wild Persian word-level lipreading dataset, comprising $743$ target words and over $414{,}000$ video samples extracted from more than $1{,}900$ hours of footage across $67$ television programs. Designed as a benchmark-ready resource, LRW-Persian provides speaker-disjoint training and test splits, wide regional and dialectal coverage, and rich per-clip metadata including head pose, age, and gender.
To ensure large-scale data quality, we establish a fully automated end-to-end curation pipeline encompassing transcription based on Automatic Speech Recognition(ASR), active-speaker localization, quality filtering, and pose/mask screening. 
We further fine-tune two widely used lipreading architectures on LRW-Persian, establishing reference performance and demonstrating the difficulty of Persian visual speech recognition. 
By filling a critical gap in low-resource languages, LRW-Persian enables rigorous benchmarking, supports cross-lingual transfer, and provides a foundation for advancing multimodal speech research in underrepresented linguistic contexts. The dataset is publicly available at: \url{https://lrw-persian.vercel.app}.
\end{abstract}

\section{Introduction}
Lip-reading, the task of inferring speech content from a silent video of a speaker’s mouth region, has attracted growing interest due to its applications in robust speech recognition under noisy conditions and assistive technologies for individuals with hearing impairments. Recent advances in deep learning, driven by large-scale, word-level benchmarks such as LRW for English \cite{chung2017lip} and LRW-1000 for Mandarin \cite{yang2019lrw}, have demonstrated that convolutional and recurrent architectures can achieve human-competitive accuracy when trained on hundreds of thousands of in-the-wild samples.

Although Persian is spoken by more than $100$ million people across several countries, visual speech research in this language remains severely under-resourced. Existing corpora either cover only small vocabularies collected under controlled conditions~\cite{hedayatipour2021pavid,gholipour2024automatic}, focus on sentence-level annotations without isolated-word segments~\cite{peymanfard2024multi}, or provide limited lexicons (e.g., around $500$ words) with restricted speaker and environmental variability~\cite{peymanfard2022word}. These limitations hinder the establishment of robust benchmarks and constrain the transfer of insights and methodologies from high-resource languages to Persian.

To address this gap, we present \emph{LRW-Persian}, a large-vocabulary, in-the-wild, word-level lip-reading corpus for Persian. LRW-Persian comprises $743$ target words extracted from over $1985$ hours of television broadcasts across multiple channels and regions of Iran, ensuring broad dialectal and contextual coverage. From this footage, we have extracted more than $414,300$ word-aligned video clips. 
A multi‐stage pipeline vets each word‐level clip to ensure high-quality, frontal, and unoccluded faces. First, the VOSK model \cite{alphacephei2020vosk} generates a transcription, and the TalkNet model \cite{beliaev2020talknet} identifies the active speaker. Next, MediaPipe \cite{lugaresi2019mediapipe} produces head-pose angles and blend-shape coefficients, while a MobileNetV2 mask detector \cite{Chandrikadeb72025FaceMask} flags potential occlusions. Finally, the DeepFace model \cite{taigman2014deepface} evaluates age, gender, and face confidence, discarding frames with low confidence, significant occlusion, or extreme head poses. The remaining segments include rich metadata for robust downstream audiovisual modeling.

Our key contributions are as follows. First, we release a publicly available Persian lip-reading dataset with a large vocabulary, balanced speaker demographics, and per-clip annotations. Second, we detail our reproducible data collection and pruning pipeline, which ensures sample quality and diversity. Third, we provide the first benchmark evaluation on this dataset by fine‐tuning two state‐of‐the‐art lip‐reading architectures, Multi‐Scale TCN \cite{martinez2020lipreading} and ResNet + BiLSTM \cite{stafylakis2017combining}, thereby establishing reference performance metrics for future research on Persian visual speech recognition.

The rest of this paper is organized as follows. Section \ref{sec: related work} surveys related work in word-level lip-reading datasets. Section \ref{sec: construction} describes our dataset construction pipeline. Section \ref{sec: stat} presents dataset statistics. Section \ref{sec: exp} details our experimental setup and baseline results. Finally, Section \ref{sec: conclusion} concludes and outlines future directions.

\section{Related Work} \label{sec: related work}
In this section, we review existing word-level lip-reading datasets across languages and then focus on Persian resources, highlighting their limitations and motivating the need for a more comprehensive Persian word-level corpus.

Various word‐level lipreading datasets have been developed for multiple languages. The Lip Reading in the Wild (LRW) dataset contains $500$ English word classes with up to $1,000$ clips per class, each $29$ frames ($1.16$ seconds) long and extracted from BBC television broadcasts under unconstrained conditions \cite{chung2017lip}. LRW-$1000$ extends this to Mandarin, offering $1,000$ syllable classes and $718,018$ samples from over $2,000$ speakers \cite{yang2019lrw}. The GRID corpus comprises $1,000$ syntactically uniform English sentences recorded under controlled conditions from $34$ speakers, with precise word‐level alignments \cite{cooke2006grid}. OuluVS2 provides multi‐view recordings of digits, phrases, and sentences from more than $50$ speakers across five viewpoints \cite{anina2015ouluvs2}. The GLips dataset consists of $250,000$ publicly available videos of the faces of the Hessian parliament German speakers, which were processed for word-level lip-reading using an automatic pipeline, formatted to match LRW for cross‐language experiments \cite{schwiebert2022multimodal}.

Recent efforts have targeted under‐resourced languages. The CLRW corpus includes $800$ Cantonese word classes with $400,000$ in‐the‐wild samples \cite{xiao2022lip}. LRW-AR is the first large‐scale Modern Standard Arabic word‐level dataset, containing $20,000$ videos across $100$ word classes \cite{daou2025cross}. A Quranic Arabic dataset provides $10,490$ multi‐view videos of single letters and Quranic words from the Al-Qaida Al-Noorania primer \cite{aljohani2023visual}. The LRWR benchmark covers $235$ Russian word classes with $117,500$ samples from $135$ speakers \cite{egorov2021lrwr}. The VLRT corpus compiles several thousand Turkish word‐level clips sourced from TV series, vlogs, and music videos \cite{berkol2022visual}. LipBengal, the largest Bengali lipreading resource, includes video data from $150$ speakers across $73$ classes \cite{sahed2025lipbengal}.

In Persian, early resources like PAVID-CV compile consonant-vowel syllables from $40$ speakers in a studio setting, limiting application to isolated syllables rather than full words \cite{hedayatipour2021pavid}. \cite{gholipour2024automatic} recorded $50$ participants uttering $25$ Persian words four times each, achieving up to $96.2\%$ accuracy with CNN-Transformer models on their custom robot-deployed dataset \cite{gholipour2024automatic}. \cite{peymanfard2022word} released the first in-the-wild Persian word-level lipreading corpus with $244,000$ clips spanning $500$ distinct words from about $1,800$ speakers; yet with only $500$ words, its lexicon remains limited \cite{peymanfard2022word}. Most recently, the Arman-AV corpus provides nearly $220$ hours of sentence-level recordings from $1,760$ speakers, but lacks isolated-word annotations \cite{peymanfard2024multi}.

Compared to previous work, our proposed dataset expands the scale and diversity of Persian lip-reading resources. It contains a significantly larger vocabulary of $743$ words and over $414,000$ high-quality samples, nearly twice the size of the largest existing lip-reading dataset. In addition to its scale, the corpus captures a wide range of regional accents and dialects, offering a more realistic representation of natural Persian speech. Unlike most prior datasets, it is fully and publicly available, enabling transparent benchmarking and reproducible research. The dataset’s breadth in vocabulary, speaker diversity, and recording conditions establishes it as the most comprehensive resource to date for advancing word-level lip-reading in Persian.

\section{Dataset Construction} \label{sec: construction}
\begin{figure}[t]
    \centering
    \includegraphics[width=0.75\linewidth]{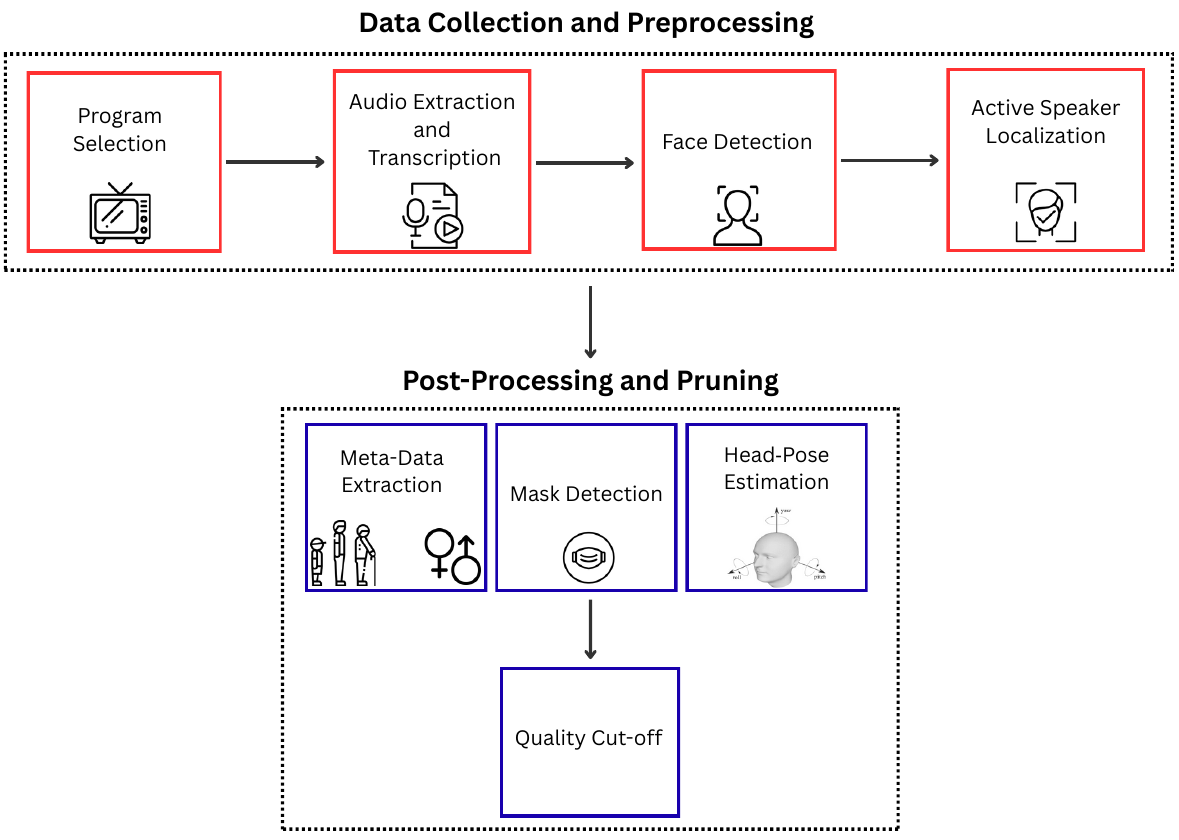}
    \caption{The end-to-end data construction pipeline for LRW-Persian.}
    \label{fig: pipeline}
\end{figure}

In this section, we detail the end-to-end pipeline for constructing our Persian lip-reading dataset, as shown in Figure \ref{fig: pipeline}. 

\subsection{Program Selection and Data Collection}
We collected the LRW-Persian dataset exclusively from Persian-language television broadcasts, encompassing a diverse range of programs, from national news bulletins and in-depth discussions on different topics to variety shows and international Persian-language channels. To ensure broad dialectal, demographic, and visual diversity (e.g., different hijab styles and ethnic backgrounds), we sourced footage from both national and regional broadcasters across multiple provinces of Iran, as well as international Persian channels. Given the significant regional variation in Persian accents, our selection included local programs produced specifically for distinct regions, thereby capturing natural linguistic and cultural variation. In total, the corpus spans $67$ programs and approximately $1{,}989$ hours of raw video. A complete list of programs is provided in Appendix~\ref{program-list}.

The video streams were retrieved from the archives of national and international broadcasters during recent years. This collection strategy ensures comprehensive coverage of frequently used Persian vocabulary and yields a naturally diverse dataset, well-suited for downstream lipreading and speech-recognition research.

\subsection{Annotations and Word Selection}

We first extracted the audio from each video using \texttt{pydub.AudioSegment} \cite{jiaaro2014pydub}, then performed offline speech recognition with the VOSK toolkit \cite{alphacephei2020vosk}, an open-source system that supports offline speech recognition for more than $20$ languages, to obtain word-level transcriptions with timing and confidence metadata. VOSK outputs recognized words and sentence segments, each tagged with start and end times and a per-word confidence score.  To ensure transcription accuracy, only words of at least four characters and with a confidence score higher than $0.9$ were retained; shorter words were excluded due to higher homophone ambiguity. For example  

From a randomly sampled $10$-hour segment of each program, we collected the $2500$ most frequent words, then intersected these lists across all channels to form $900$ candidate targets.  A subsequent manual review pruned this set to $743$ words that had enough number of high-quality samples.

Next, we searched the entire corpus for every occurrence of our $743$-word list, retaining only the resulting video clips shorter than $1.5$ s to maximize transcription correctness and reduce computational cost.  This procedure yielded a collection of word-level video segments, each annotated with its precise time interval.  

\subsection{Face Detection and Active‐Speaker Localization}
For face detection and active-speaker localization, we employed the TalkNet framework \cite{beliaev2020talknet}, an end-to-end audio-visual framework that captures both short-term and long-term temporal dependencies via self-attention and cross-attention layers, to identify, among multiple on-screen faces, which track corresponds to the speaking subject.  Faces were first detected per frame, linked over time into continuous tracks, and then scored by TalkNet to select the active speaker in each clip.  The output for each word consists of a tightly cropped face‐track showing the speaker articulating that word. To ensure minimal visual quality, we then discarded any clips whose spatial resolution fell below $100\times100$ pixels.

\subsection{Post-Processing and Pruning}
To ensure that all word-level clips are high-quality, frontal, and unobstructed, we applied a three-stage post-processing pipeline. 
First, we employed two pre-trained models: the MediaPipe FaceLandmarker~\cite{lugaresi2019mediapipe} to extract 3D head poses and blend-shape coefficients, and a MobileNetV2-based mask detector~\cite{Chandrikadeb72025FaceMask}. Helper functions were used to convert rotation matrices into roll, pitch, and yaw angles. 
Second, for each frame, we computed head-pose angles and blend-shape coefficients using MediaPipe to capture subtle facial dynamics. 
Third, we utilized DeepFace~\cite{taigman2014deepface} to estimate age, gender, and face-confidence scores, discarding samples with confidence values below $0.75$. 

Because part of the dataset originates from television archives recorded during the COVID-19 pandemic, some speakers appeared wearing face masks, rendering lip movements invisible. To address this, we filtered out frames with a detected mask probability greater than $0.75$. We further discarded the she samples with extreme head poses satisfying $|yaw| > 30^\circ$ or $|pitch| > 40^\circ$. 

\begin{figure}[t!]
    \centering
    \includegraphics[width=0.8\linewidth]{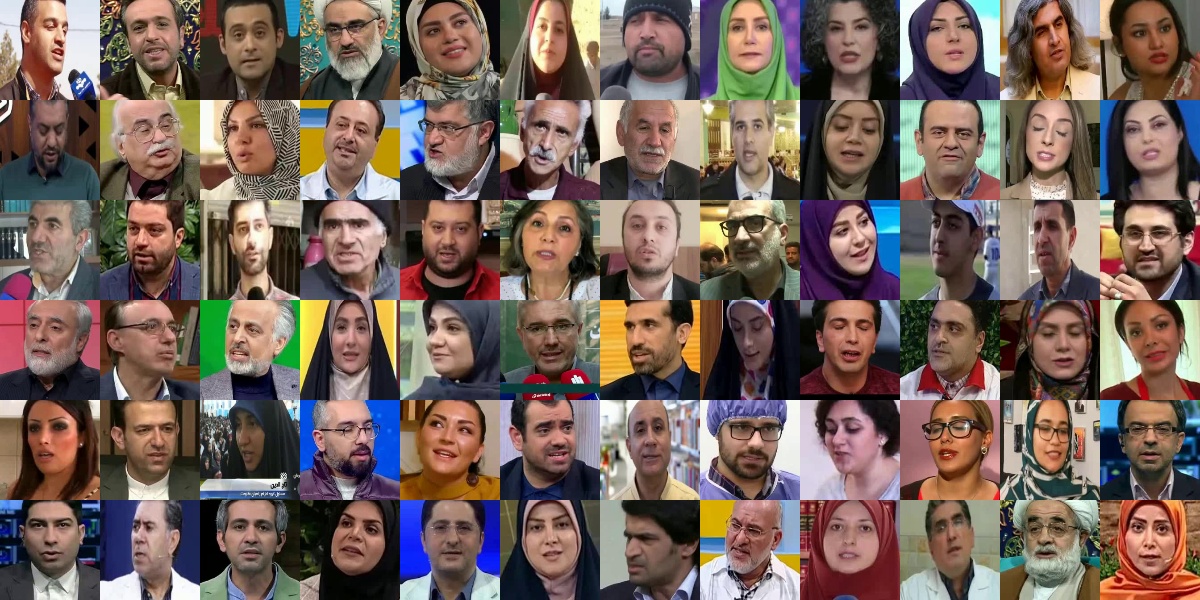}
    \caption{Example samples from the LRW-Persian dataset.
    Representative frames illustrating diversity in speakers, lighting conditions, and head poses across different programs.}
    \label{fig:samples}
\end{figure}

The resulting metadata, including head pose and demographic attributes, is stored and is provided along with the samples in the dataset to facilitate downstream analysis and model training. Figure~\ref{fig:samples} illustrates the diversity of speakers, poses, and recording conditions captured in the dataset.

Representative samples from the LRW-Persian dataset are shown in Figure~\ref{fig:samples}, illustrating the diversity in speakers, poses, and recording conditions.

\section{Dataset Statistics} \label{sec: stat}
This section provides an overview of the video corpus and the derived samples in our lipreading dataset.

The source corpus comprises $1{,}989$ hours of footage across $2{,}480$ raw video files drawn from $67$ television programs listed in Appendix~\ref{program-list}. In total, $8{,}082{,}286$ raw word tokens were automatically recognized from the source videos. After preprocessing, this number was reduced to $493{,}603$ tokens, and subsequent post-processing yielded $414{,}308$ high-quality word-level samples.

\begin{figure}[H]
    \centering
    \includegraphics[width=0.6\linewidth]{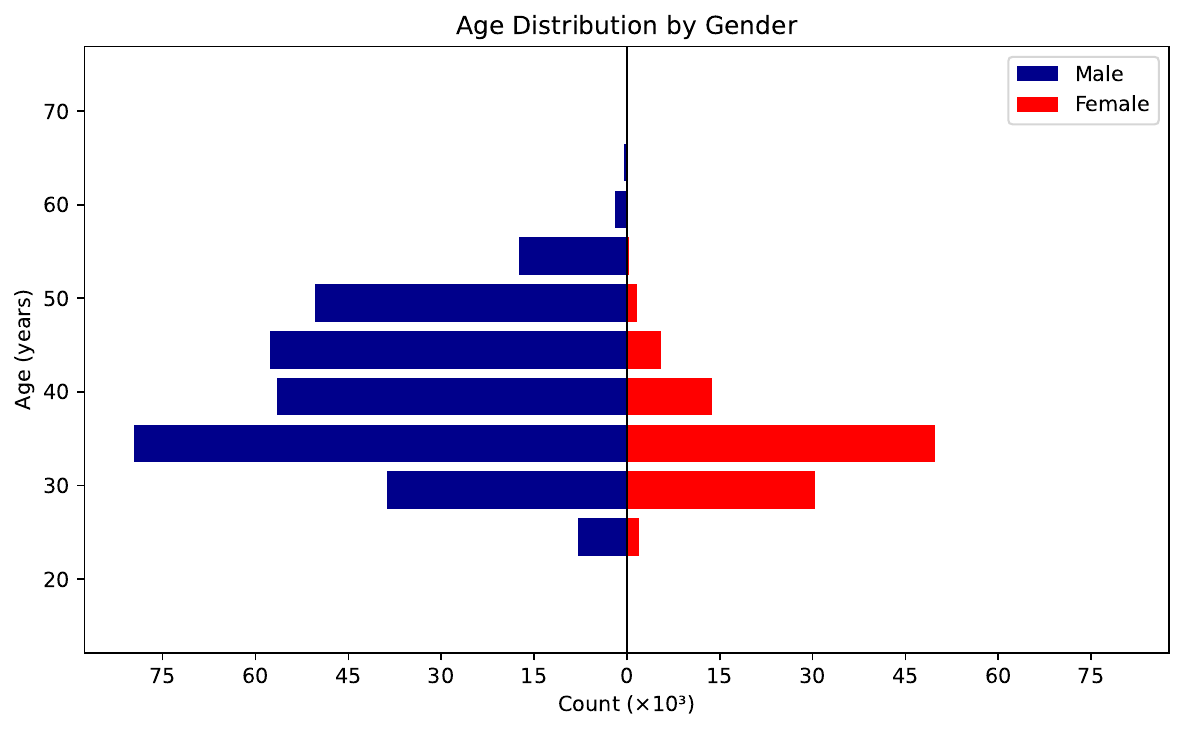}
    \caption{Population pyramid illustrating the age distribution of male (blue) and female (red) speakers, with horizontal bars representing the number of speakers (in thousands) across 5-year age bins.}
    \label{fig:age_gender}
\end{figure}

The dataset exhibits a gender distribution of approximately $75\%$ male and $25\%$ female. Figure~\ref{fig:age_gender} presents the age distribution in $5$-year intervals, showing balanced representation across a wide range of age groups.

\begin{figure}[h]
    \centering
    \includegraphics[width=0.7\linewidth]{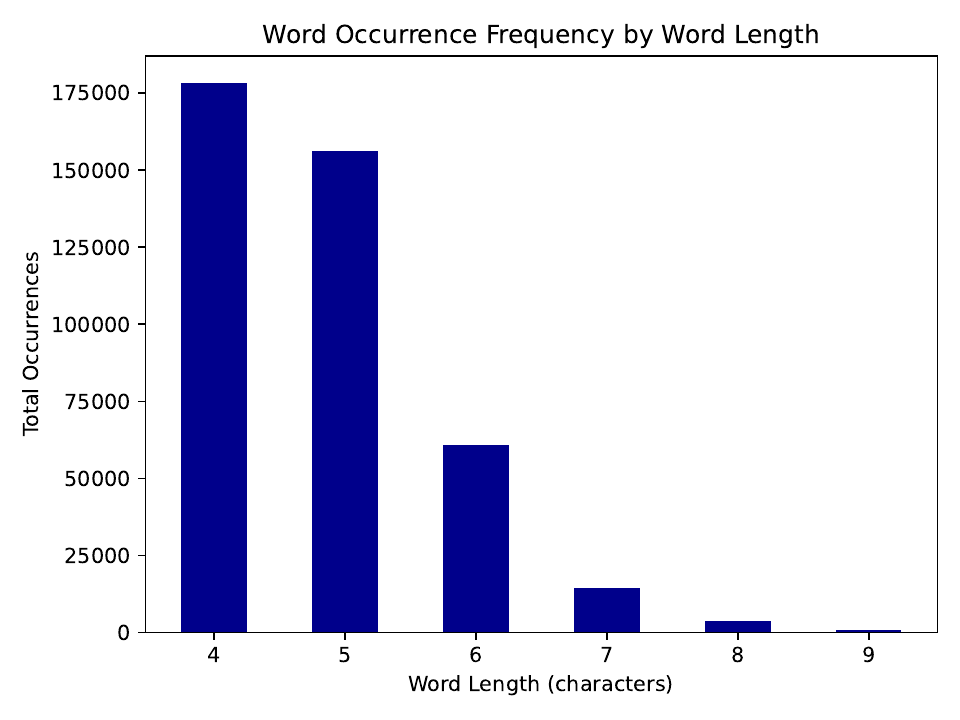}
    \caption{Histogram showing the frequency of each word length (in characters) across all samples in the dataset.}
    \label{fig:word_char}
\end{figure}

At the lexical level, Figure~\ref{fig:word_char} shows the distribution of word lengths (in characters). The histogram peaks at four-character words, with frequencies gradually decreasing for longer tokens.

\begin{figure}[h]
    \centering
    \includegraphics[width=0.3\linewidth]{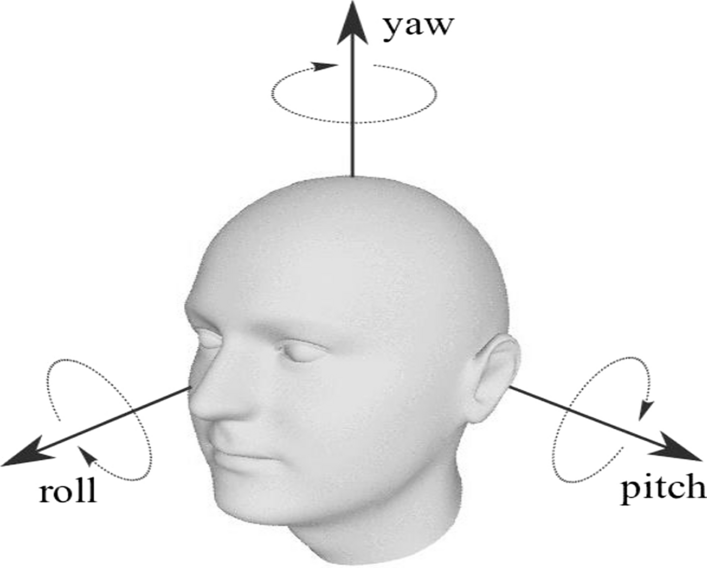}
    \caption{Illustration of roll, pitch, and yaw. Figure from \cite{dubey2022image}.}
    \label{fig:euler-angles}
\end{figure}

\begin{figure}[t]
  \centering
  \begin{subfigure}{0.5\columnwidth}
    \includegraphics[width=\linewidth]{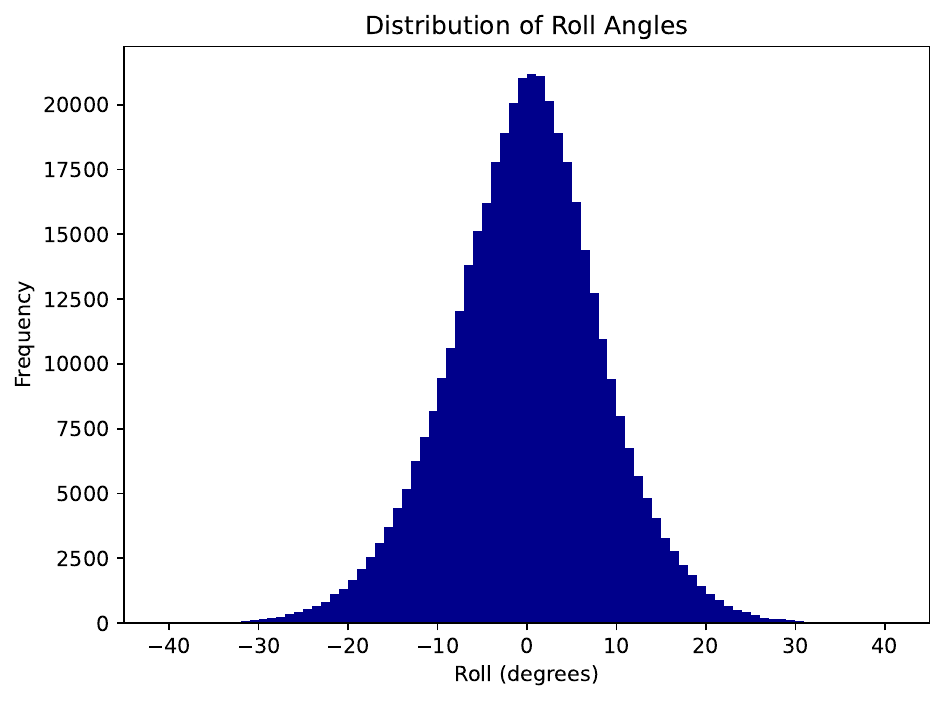}
    \label{fig:roll_dist}
  \end{subfigure}\hfill
  \begin{subfigure}{0.5\columnwidth}
    \includegraphics[width=\linewidth]{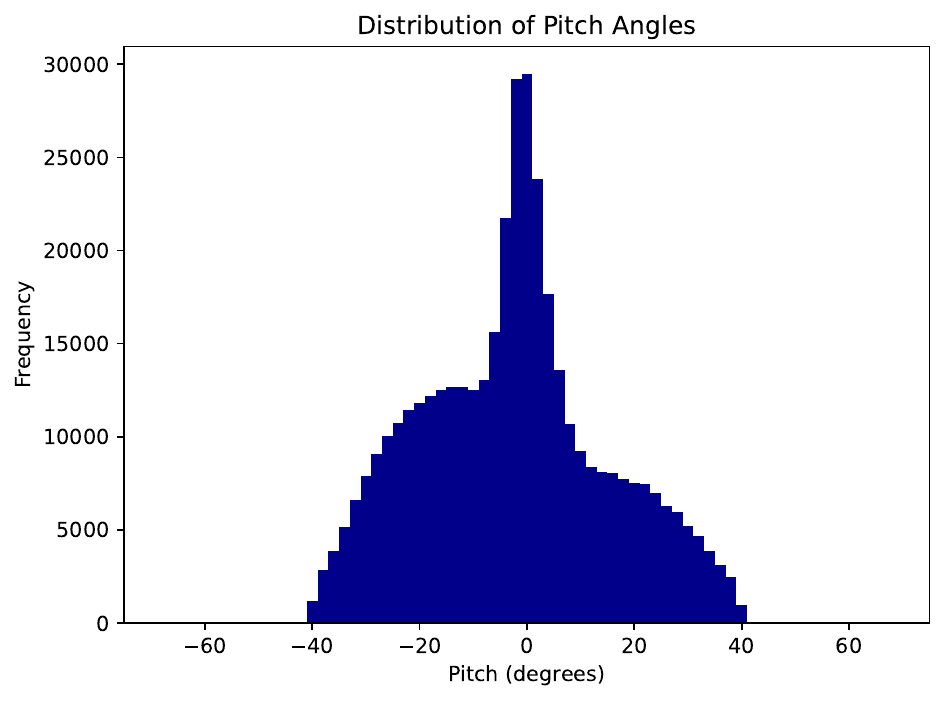}
    \label{fig:pitch_dist}
  \end{subfigure}\hfill
  \begin{subfigure}{0.5\columnwidth}
    \includegraphics[width=\linewidth]{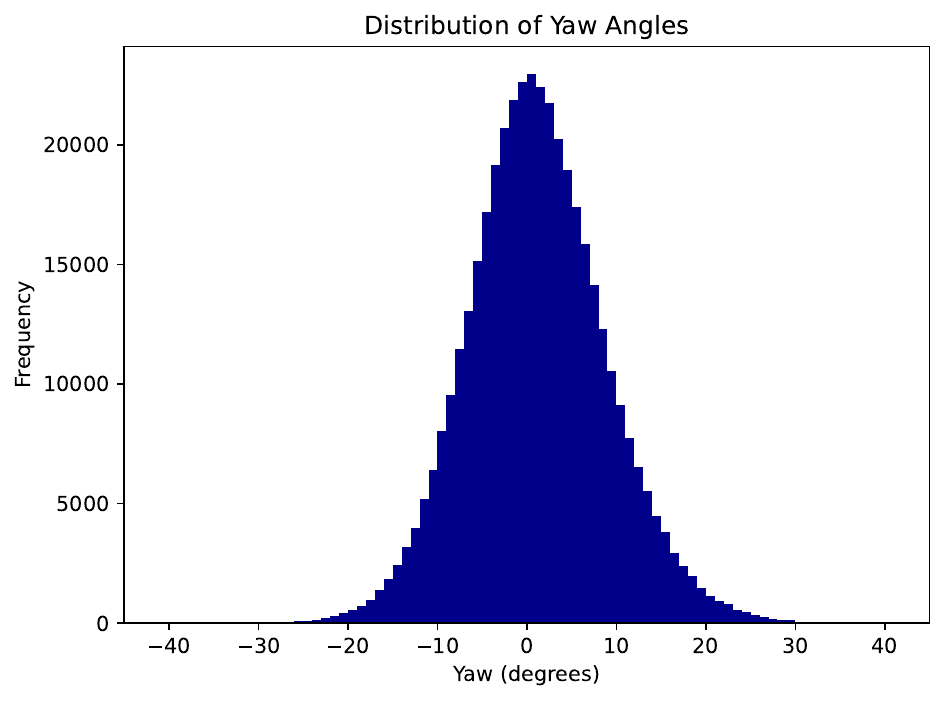}
    \label{fig:yaw_dist}
  \end{subfigure}
  \caption{Frequency histograms of roll, pitch, and yaw angles for all the samples in the dataset.}
  \label{fig:angle_distributions}
\end{figure}

Head-pose variability was analyzed using the roll, pitch, and yaw angles estimated for each frame. Figure~\ref{fig:euler-angles} illustrates these Euler angles, while Figure~\ref{fig:angle_distributions} shows their frequency distributions across the dataset. The yaw and roll distributions are tightly centered around $0^{\circ}$, indicating that most samples feature near-frontal faces, with frequencies rapidly decreasing as angles increase. In contrast, the pitch distribution exhibits greater variability: although most frames cluster within $\pm10^{\circ}$, longer tails extend toward the $\pm40^{\circ}$ detection thresholds, our filtering limits, reflecting occasional upward or downward head movements. This natural variation in head orientation ensures that the dataset captures realistic, in-the-wild appearance diversity, which is essential for developing lipreading models that generalize robustly to unconstrained scenarios.

Finally, the dataset is partitioned into disjoint training and test subsets. The split proportions are Train: $78\%$ ($324{,}405$ samples) and Test: $22\%$ ($89{,}903$ samples). To prevent speaker overlap and data leakage, the programs assigned to the training and test sets are completely distinct, ensuring that no speaker or video segment appears in both splits.


\section{Experiments} \label{sec: exp}
This section presents the evaluation of existing lipreading models on the proposed dataset. 
We assess the performance of two established architectures widely recognized in visual speech recognition for their ability to capture spatiotemporal dynamics from mouth-region video sequences: the Multi-Scale Temporal Convolutional Network (Multi-Scale TCN)~\cite{martinez2020lipreading} and the Residual Network combined with Bidirectional LSTMs (ResNet + BiLSTM)~\cite{stafylakis2017combining}.

\textbf{Multi-Scale TCN.} 
The Multi-Scale Temporal Convolutional Network (TCN)~\cite{martinez2020lipreading} employs temporal convolutions to model sequential dependencies in visual speech. The architecture is based on a ResNet-18 backbone, incorporating an initial 3D convolutional layer for spatiotemporal feature extraction, followed by a stack of multi-scale temporal convolutional blocks designed to capture both short-term and long-term temporal patterns.

\textbf{ResNet + BiLSTM.} 
The model proposed by~\cite{stafylakis2017combining} integrates spatial and temporal modeling through a combination of a 3D convolutional front-end, a 34-layer Residual Network, and a two-layer Bidirectional LSTM. The 3D convolutional module captures short-term lip motion dynamics, while the ResNet extracts robust per-frame spatial representations. These features are then fed into the BiLSTM, which learns long-range temporal dependencies across the input sequences.

The subsequent sections describe the training configurations and quantitative results obtained for both architectures on the LRW-Persian dataset.

\subsection{Experimental Setup}

\textbf{Training Details.}  
Both models are fine-tuned for $30$ epochs on an NVIDIA RTX~$3090$ GPU using the AdamW optimizer with a learning rate of \(3 \times 10^{-4}\).  
For the Multi-Scale TCN, we use a batch size of $32$ and a weight decay of \(1 \times 10^{-2}\).  
For the ResNet + BiLSTM model, we use a batch size of $64$ and a weight decay of \(1 \times 10^{-4}\).  

\textbf{Evaluation Metrics.}  
To evaluate recognition performance, we report Top-$k$ accuracies, where Top-1 corresponds to the percentage of correctly predicted words, and Top-$k$ represents the percentage of samples for which the correct word appears among the model’s top $k$ predictions.  
This metric provides a comprehensive measure of performance in large-vocabulary, multi-class settings, where visually similar words may exhibit overlapping viseme patterns.

\subsection{Results}
Table~\ref{tab:performance} summarizes the quantitative performance of both models on the test set. The Multi‐Scale TCN model achieves a higher overall accuracy, benefiting from its efficient temporal modeling and multi‐scale feature aggregation, while the ResNet + BiLSTM model performs competitively, highlighting its capacity for capturing sequential dependencies through recurrent connections.

These results demonstrate that both models achieve lower performance compared to their counterparts trained on standard English lipreading datasets such as LRW~\cite{martinez2020lipreading,stafylakis2017combining}. This performance gap highlights the inherent difficulty of the proposed Persian lipreading corpus, which features a diverse set of speakers, natural and unconstrained lighting conditions, and fine-grained visual distinctions between phonetically similar words. These characteristics make accurate recognition particularly challenging. 

The dataset’s complexity thus establishes it as a valuable benchmark for evaluating and advancing future lipreading models, especially in low-resource language settings.

\begin{table}[ht]
  \centering
  \caption{Performance of the baseline model on our Persian lip‐reading corpus.}
  \label{tab:performance}
  \resizebox{\textwidth}{!}{
  \begin{tabular}{lccc}
    \toprule
    Model       & Top-1 Accuracy (\%) & Top-5 Accuracy (\%)  &  Top-10 Accuracy(\%) \\
    \midrule
    Multi-Scale TCN~\cite{martinez2020lipreading}  &           41.09       &         65.53   &     72.99    \\
    ResNet + BiLSTM ~\cite{stafylakis2017combining} &           34.09       &         58.70   &     67.08  \\
    \bottomrule
  \end{tabular}
  }
\end{table}

\section{Conclusion} \label{sec: conclusion}

In this work, we introduced \emph{LRW-Persian}, the largest in-the-wild Persian word-level lip-reading corpus to date. Spanning $743$ target words and $414,308$ high-quality, face-aligned video clips extracted from $1,989$ hours of broadcast footage across $67$ programs. LRW-Persian represents a significant step toward large-scale visual speech recognition in Persian. The dataset includes per-clip metadata, such as head-pose angles, age, and gender, enabling a wide range of audiovisual modeling and analysis tasks. Our end-to-end data curation pipeline, integrating VOSK transcription, TalkNet-based speaker localization, MediaPipe, and DeepFace filtering, ensures that only clean, frontal, and unoccluded samples are retained.

In our experimental evaluation, we evaluated two state-of-the-art lip-reading architectures, the Multi-Scale Temporal Convolutional Network (Multi-Scale TCN) and the Residual Network combined with Bidirectional LSTMs (ResNet + BiLSTM), on the LRW-Persian dataset. 
The Multi-Scale TCN achieved superior overall accuracy, highlighting its ability to capture multi-scale temporal dependencies effectively, while the ResNet + BiLSTM exhibited competitive performance through its recurrent temporal modeling. These results underscore LRW-Persian’s value as a challenging and comprehensive benchmark for Persian lip-reading, revealing particular areas for improvement, such as robustness to extreme head poses and handling of low-frequency word classes.

By filling a critical gap in Persian visual-speech resources, LRW-Persian establishes a strong foundation for advancing research in low-resource languages. It also facilitates cross-lingual transfer learning and the development of multi-modal speech technologies tailored to Persian speakers.

\newpage

\bibliography{references} 

\newcommand{\etalchar}[1]{$^{#1}$}
\begin{thebibliography}{DBHRK25}

\bibitem[AJ23]{aljohani2023visual}
Nada~Faisal Aljohani and Emad~Sami Jaha.
\newblock Visual lip-reading for quranic arabic alphabets and words using deep learning.
\newblock {\em Computer Systems Science \& Engineering}, 46(3), 2023.

\bibitem[{Alp}20]{alphacephei2020vosk}
{Alpha Cephei}.
\newblock Vosk: Offline speech recognition toolkit.
\newblock \url{https://alphacephei.com/vosk/}, 2020.
\newblock Accessed: 2025-07-21.

\bibitem[AZZP15]{anina2015ouluvs2}
Iryna Anina, Ziheng Zhou, Guoying Zhao, and Matti Pietik{\"a}inen.
\newblock Ouluvs2: A multi-view audiovisual database for non-rigid mouth motion analysis.
\newblock In {\em 2015 11th IEEE international conference and workshops on automatic face and gesture recognition (FG)}, volume~1, pages 1--5. IEEE, 2015.

\bibitem[B{\c{C}}E22]{berkol2022visual}
A~Berkol, M~{\c{C}}olak, and H~Erdem.
\newblock Visual lip reading dataset in turkish. data 8 (1): 15, 2022.

\bibitem[BRG20]{beliaev2020talknet}
Stanislav Beliaev, Yurii Rebryk, and Boris Ginsburg.
\newblock Talknet: Fully-convolutional non-autoregressive speech synthesis model.
\newblock {\em arXiv preprint arXiv:2005.05514}, 2020.

\bibitem[CBCS06]{cooke2006grid}
Martin Cooke, Jon Barker, Stuart Cunningham, and Xu~Shao.
\newblock The grid audio-visual speech corpus.
\newblock {\em Zenodo}, 2006.

\bibitem[CZ17]{chung2017lip}
Joon~Son Chung and Andrew Zisserman.
\newblock Lip reading in the wild.
\newblock In {\em Computer Vision--ACCV 2016: 13th Asian Conference on Computer Vision, Taipei, Taiwan, November 20-24, 2016, Revised Selected Papers, Part II 13}, pages 87--103. Springer, 2017.

\bibitem[DBHRK25]{daou2025cross}
Samar Daou, Achraf Ben-Hamadou, Ahmed Rekik, and Abdelaziz Kallel.
\newblock Cross-attention fusion of visual and geometric features for large-vocabulary arabic lipreading.
\newblock {\em Technologies}, 13(1):26, 2025.

\bibitem[Deb]{Chandrikadeb72025FaceMask}
Chandrika Deb.
\newblock Face mask detection.
\newblock In {\em GitHub Repository}.
\newblock [Online]. Available: \url{https://github.com/chandrikadeb7/Face-Mask-Detection}.

\bibitem[DT22]{dubey2022image}
Deepika Dubey and Geetam~Singh Tomar.
\newblock Image alignment in pose variations of human faces by using corner detection method and its application for pifr system.
\newblock {\em Wireless Personal Communications}, 124(1):147--162, 2022.

\bibitem[EKKK21]{egorov2021lrwr}
Evgeniy Egorov, Vasily Kostyumov, Mikhail Konyk, and Sergey Kolesnikov.
\newblock Lrwr: large-scale benchmark for lip reading in russian language.
\newblock {\em arXiv preprint arXiv:2109.06692}, 2021.

\bibitem[GMGT24]{gholipour2024automatic}
Amir Gholipour, Hoda Mohammadzade, Ali Ghadami, and Alireza Taheri.
\newblock Automatic lip reading of persian words by a robotic system using deep learning algorithms.
\newblock {\em Iranian Journal of Science and Technology, Transactions of Electrical Engineering}, 48(4):1519--1538, 2024.

\bibitem[HSM21]{hedayatipour2021pavid}
Mahsa Hedayatipour, Yasser Shekofteh, and Mohsen~Ebrahimi Moghaddam.
\newblock Pavid-cv s: Persian audio-visual database of cv syllables.
\newblock In {\em 2021 29th Iranian Conference on Electrical Engineering (ICEE)}, pages 470--473. IEEE, 2021.

\bibitem[LTN{\etalchar{+}}19]{lugaresi2019mediapipe}
Camillo Lugaresi, Jiuqiang Tang, Hadon Nash, Chris McClanahan, Esha Uboweja, Michael Hays, Fan Zhang, Chuo-Ling Chang, Ming Yong, Juhyun Lee, et~al.
\newblock Mediapipe: A framework for perceiving and processing reality.
\newblock In {\em Third workshop on computer vision for AR/VR at IEEE computer vision and pattern recognition (CVPR)}, volume 2019, 2019.

\bibitem[MMPP20]{martinez2020lipreading}
Brais Martinez, Pingchuan Ma, Stavros Petridis, and Maja Pantic.
\newblock Lipreading using temporal convolutional networks.
\newblock In {\em ICASSP 2020-2020 IEEE International Conference on Acoustics, Speech and Signal Processing (ICASSP)}, pages 6319--6323. IEEE, 2020.

\bibitem[PHL{\etalchar{+}}24]{peymanfard2024multi}
Javad Peymanfard, Samin Heydarian, Ali Lashini, Hossein Zeinali, Mohammad~Reza Mohammadi, and Nasser Mozayani.
\newblock A multi-purpose audio-visual corpus for multi-modal persian speech recognition: The arman-av dataset.
\newblock {\em Expert Systems with Applications}, 238:121648, 2024.

\bibitem[PLH{\etalchar{+}}22]{peymanfard2022word}
Javad Peymanfard, Ali Lashini, Samin Heydarian, Hossein Zeinali, and Nasser Mozayani.
\newblock Word-level persian lipreading dataset.
\newblock In {\em 2022 12th International Conference on Computer and Knowledge Engineering (ICCKE)}, pages 225--230. IEEE, 2022.

\bibitem[Rob14]{jiaaro2014pydub}
James Robert.
\newblock pydub: Manipulate audio with a simple and easy high level interface.
\newblock \url{https://github.com/jiaaro/pydub}, 2014.
\newblock Accessed: 2025-07-21.

\bibitem[SAN{\etalchar{+}}25]{sahed2025lipbengal}
Md~Tanvir~Rahman Sahed, Md~Tanjil~Islam Aronno, Hussain Nyeem, Md~Abdul Wahed, Tashrif Ahsan, R~Rafiul Islam, Tareque~Bashar Ovi, Manab~Kumar Kundu, and Jane~Alam Sadeef.
\newblock Lipbengal: Pioneering bengali lip-reading dataset for pronunciation mapping through lip gestures.
\newblock {\em Data in Brief}, 58:111254, 2025.

\bibitem[ST17]{stafylakis2017combining}
Themos Stafylakis and Georgios Tzimiropoulos.
\newblock Combining residual networks with lstms for lipreading.
\newblock {\em arXiv preprint arXiv:1703.04105}, 2017.

\bibitem[SWQ{\etalchar{+}}22]{schwiebert2022multimodal}
Gerald Schwiebert, Cornelius Weber, Leyuan Qu, Henrique Siqueira, and Stefan Wermter.
\newblock A multimodal german dataset for automatic lip reading systems and transfer learning.
\newblock {\em arXiv preprint arXiv:2202.13403}, 2022.

\bibitem[TYRW14]{taigman2014deepface}
Yaniv Taigman, Ming Yang, Marc'Aurelio Ranzato, and Lior Wolf.
\newblock Deepface: Closing the gap to human-level performance in face verification.
\newblock In {\em Proceedings of the IEEE conference on computer vision and pattern recognition}, pages 1701--1708, 2014.

\bibitem[XTZ{\etalchar{+}}22]{xiao2022lip}
Yewei Xiao, Lianwei Teng, Aosu Zhu, Xuanming Liu, and Picheng Tian.
\newblock Lip reading in cantonese.
\newblock {\em IEEE Access}, 10:95020--95029, 2022.

\bibitem[YZF{\etalchar{+}}19]{yang2019lrw}
Shuang Yang, Yuanhang Zhang, Dalu Feng, Mingmin Yang, Chenhao Wang, Jingyun Xiao, Keyu Long, Shiguang Shan, and Xilin Chen.
\newblock Lrw-1000: A naturally-distributed large-scale benchmark for lip reading in the wild.
\newblock In {\em 2019 14th IEEE international conference on automatic face \& gesture recognition (FG 2019)}, pages 1--8. IEEE, 2019.

\end{thebibliography}
\bibliographystyle{alpha}

\newpage

\begin{appendices}

\section{List of Programs} \label{program-list}

Table \ref{tab:program-categories} shows the full list of $67$ television programs that are used to collect the source videos.

{\centering
\begin{longtable}{@{}l
                  >{\centering\arraybackslash}p{0.7cm}
                  >{\centering\arraybackslash}p{0.7cm}
                  >{\centering\arraybackslash}p{0.7cm}
                  >{\centering\arraybackslash}p{0.7cm}
                  >{\centering\arraybackslash}p{0.7cm}
                  c@{}}
\caption{Source programs and their categories.}
\label{tab:program-categories}\\
\toprule
\textbf{Program}
  & \rheader{\textbf{News}}
  & \rheader{\textbf{Entertainment}}
  & \rheader{\textbf{Discussion}}
  & \rheader{\textbf{Lifestyle}}
  & \rheader{\textbf{Documentary}}
  & \rheader{\textbf{Total Hours}}\\
\midrule
\endfirsthead

\toprule
\textbf{Program}
  & \rheader{\textbf{News}}
  & \rheader{\textbf{Entertainment}}
  & \rheader{\textbf{Discussion}}
  & \rheader{\textbf{Lifestyle}}
  & \rheader{\textbf{Documentary}}
  & \rheader{\textbf{Total Hours}}\\
\midrule
\endhead
    News-20:30 &  \checkmark     &               &                  &    &  &    22       \\ 
    News-21:00 &  \checkmark     &               &                  &    &  &        11       \\  
    News-09:00 &  \checkmark     &               &                  &       &  &     8       \\      
    Sarasari &   \checkmark    &               &                  &      &  &     26        \\
    Nimroozi &  \checkmark     &               &                  &      &  &      38       \\    
    Isfahan-23:00 &  \checkmark     &               &                  &     &  &      34        \\
    Isfahan-20:00 &  \checkmark     &               &                  &    &  &       47        \\
    Razavi-20:45 &  \checkmark     &               &                  &    &  &       43        \\
    Abadan-23:00 &  \checkmark     &               &                  &    &  &       33        \\
    Ilam-23:00 &  \checkmark     &               &                  &     &  &        54      \\    
    Simaye Khanevadeh &       &  \checkmark             &                  &    &    &        39       \\
    Salam Sobh bekheir &       &  \checkmark             &                  &    &  &        64       \\ 
    Sobh Bekheir Iran &       &     \checkmark          &                  &     &  &       69       \\   
    Sobh-e-Aali &       &      \checkmark         &                  &   &  &        57         \\  
    Bala be door &       &     \checkmark          &                  &      &  &      19       \\
    Gheseye Shab &       &      \checkmark        &                  &    &  &        6       \\        
    Ojoobeha &      &       \checkmark         &                  &   &  &         29       \\   
    Befarmaied Sham &      &    \checkmark      &                  &    &  &         159      \\ 
    Salam Tehran &       &      \checkmark         &                  &   &  &        38        \\            
    Porseshgar &       &               &        \checkmark          &     &  &       19       \\ 
    Faramatn &       &               &       \checkmark           &     &  &      55        \\
    Jahan Ara &      &               &        \checkmark           &     &  &       42       \\  
    Tablet &       &            &        \checkmark          &    &  &         10      \\  
    Panj Dari &       &               &          \checkmark        &      &  &       13      \\  
    Keshik-e-Salamat &       &               &    \checkmark              &   &  &         20        \\  
    Harekat &       &              &    \checkmark               &     &  &       9       \\ 
    Varoonegi &       &               &        \checkmark          &     &  &      20        \\     
    Aftab-e-Sharghi &       &               &    \checkmark              &    &  &      22         \\         
    Sobhaneh Irani &       &               &   \checkmark               &   &  &        113        \\         
    Pavaraghi Ketab &       &            &       \checkmark           &    &  &        7       \\    
    Mosbat-e-ketab &       &            &          \checkmark        &      &  &      20       \\    
    Tar-o-Toranj &       &               &         \checkmark         &    &  &      30      \\   
    Hoosh-e-Masnoi &       &              &     \checkmark             &    &  &        12       \\   
    Khane Injast &       &               &          \checkmark        &  &  &         2        \\   
    Vaght-e-Andisheh &       &               &      \checkmark            &    &   &     7       \\   
    Shatranj &       &               &         \checkmark         &   &  &       33         \\    
    Charkh &       &               &         \checkmark         &     &  &       22       \\ 
    Cheragh Motalee &       &              &      \checkmark             &     &  &      8        \\ 
    Hal-e-Khoob &       &                &        \checkmark          & \checkmark &  &        48         \\
    Khooneye Khodetoone &       &              &   \checkmark               &  \checkmark  &  &       130        \\  
    Kam-e-Shirin &       &               &          \checkmark        &   \checkmark &  &       27        \\
    Madar-Koodak &      &               &         \checkmark         &  \checkmark  &  &       44        \\   
    Salam Donya &       &          &    \checkmark               &  \checkmark   &  &       43        \\   
    Shafa &      &               &        \checkmark           &   \checkmark    &  &      8       \\   
    Mosbat-e-Salamat &      &               &     \checkmark              &  \checkmark  &  &       61        \\
    Doctor Salam &      &               &       \checkmark           &  \checkmark  &  &        42       \\   
    Panjere baz &       &      \checkmark         &      \checkmark       &   \checkmark &  &      31   \\
    Dastan &       &        \checkmark       &                 \checkmark &    &  &        26       \\   
    Be Khane Barmigardim &       &      \checkmark         &      \checkmark            &  \checkmark  &  \checkmark &       125        \\   
    Karkhoone &       &               &           \checkmark       &      & \checkmark &       67      \\      
    Revayat-e-Motabar &       &              &                  &    & \checkmark  &        58       \\    
    Nan-o-Honar &       &               &                  &    &  \checkmark &       10        \\    
    Beriz Bepaz &      &                &              &  \checkmark  &  &         6      \\      

    \bottomrule
 
\end{longtable}
}

\end{appendices}

\end{document}